\documentclass[letterpaper, 10 pt, conference]{ieeeconf}  

\IEEEoverridecommandlockouts                              

\usepackage[sorting=none, style=numeric-comp, maxbibnames=99]{biblatex}

\usepackage{color}
\usepackage{epsfig}
\usepackage{graphicx}
\usepackage{algorithm,algorithmic}

\usepackage{adjustbox}
\usepackage{array}
\usepackage{booktabs}
\usepackage{colortbl}
\usepackage{float,wrapfig}
\usepackage{framed}
\usepackage{hhline}
\usepackage{multirow}
\usepackage{subcaption} 
\usepackage[font=small]{caption}
\usepackage[percent]{overpic}

\usepackage{amsmath,amsfonts,amssymb}
\usepackage{amsthm} 
\usepackage{bm}
\usepackage{nicefrac}
\usepackage{microtype}
\usepackage{contour}
\usepackage{courier}

\usepackage{changepage}
\usepackage{extramarks}
\usepackage{fancyhdr}
\usepackage{lastpage}
\usepackage{setspace}
\usepackage{soul}
\usepackage{xspace}
\usepackage{cuted}
\usepackage{fancybox}
\usepackage{afterpage}

\usepackage[breaklinks=true,colorlinks,backref=True]{hyperref}
\hypersetup{colorlinks,linkcolor={black},citecolor={CiteBlue},urlcolor={magenta}}
\usepackage{url}
\usepackage{quoting}
\usepackage{epigraph}

\usepackage{enumerate}
\usepackage{paralist,tabularx}
\usepackage{comment}
\usepackage{pdfpages}
\usepackage{caption}
\usepackage{subcaption}

\usepackage{pifont}

\usepackage{enumitem}

\makeatletter
\DeclareRobustCommand\onedot{\futurelet\@let@token\@onedot}
\def\@onedot{\ifx\@let@token.\else.\null\fi\xspace}

\def\etal{et al\onedot}

\makeatother

\definecolor{MyDarkBlue}{rgb}{0,0.08,1}
\definecolor{MyDarkGreen}{rgb}{0.02,0.6,0.02}
\definecolor{MyDarkRed}{rgb}{0.8,0.02,0.02}
\definecolor{MyDarkOrange}{rgb}{0.40,0.2,0.02}
\definecolor{MyPurple}{RGB}{111,0,255}
\definecolor{MyRed}{rgb}{1.0,0.0,0.0}
\definecolor{MyGold}{rgb}{0.75,0.6,0.12}
\definecolor{MyDarkgray}{rgb}{0.66, 0.66, 0.66}
\definecolor{CiteBlue}{rgb}{0.4, 0.55, 0.8}

\def\OURS{MORPH\xspace}

\def\NOHW{RL-NoHWOpt\xspace}
\def\TRANSFORM{Transform2Act\xspace}
\def\CMAES{CMA-ES with RL inner-loop\xspace}

\title{\LARGE \bf
MORPH: Design Co-optimization with Reinforcement Learning\\via a Differentiable Hardware Model Proxy
}

\author{Zhanpeng He and Matei Ciocarlie \\ 
\url{https://roamlab.github.io/morph/}
\thanks{Zhanpeng He is with the Department of Computer Science, Columbia University, New York, USA.
        {\tt\small zhanpeng@cs.columbia.edu}}%
\thanks{Matei Ciocarlie is with the Department of Mechanical Engineering, Columbia University, New York, USA.
        {\tt\small matei.ciocarlie@columbia.edu}}
}
\begin{document}

\maketitle
\thispagestyle{empty}
\pagestyle{empty}

\begin{abstract}
We introduce \OURS, a method for co-optimization of hardware design parameters and control policies in simulation using reinforcement learning. Like most co-optimization methods, \OURS relies on a model of the hardware being optimized, usually simulated based on the laws of physics. However, such a model is often difficult to integrate into an effective optimization routine. To address this, we introduce a proxy hardware model, which is always differentiable and enables efficient co-optimization alongside a long-horizon control policy using RL. \OURS is designed to ensure that the optimized hardware proxy remains as close as possible to its realistic counterpart, while still enabling task completion. We demonstrate our approach on simulated 2D reaching and 3D multi-fingered manipulation tasks.\end{abstract}

\section{INTRODUCTION}
\label{sec:into}

Design optimization and automation techniques generally aim to alleviate some of the time-consuming process of hardware design, usually by iterating over a large parameter space in simulation in order to find optimal (or good enough) design parameters before the hardware is ever constructed. Within this broad category, co-design or co-optimization methods use simulation in order to simultaneously optimize both hardware parameters and aspects of the software that will run on the hardware (e.g. a controller or a policy) to ensure suitability for a specific task.

Creating a simulated model for the hardware being optimized is a crucial component of design automation. In the case of co-design, the importance of this step only grows, since the simulated hardware model must not only be faithful to its real counterpart, but also lend itself to optimization techniques capable of simultaneously handling the controller or policy component of the co-optimization.

Given the recent success of reinforcement learning (RL) methods in optimizing effective control policies for complex behaviors, it is only natural for the field to apply RL techniques to the co-design problem as well. The core idea of this approach is to compute policy gradients for both the design parameters and control policy parameters. For example, one way to achieve this is by treating a design, or a change of the design, as actions, and co-learn design actions and control actions~\cite{Yuan2021Transform2ActLA}. However, this results in extending the action space in ways that can increase the difficulty of exploration. Another approach is to integrate the design parameters and their effect with the control policy via differentiable physics~\cite{chen2020hwasp}. However, this integration relies on the existence of a differentiable modeling of a task, which may not be available.

\begin{figure}
    \centering
    \includegraphics[width=\linewidth]{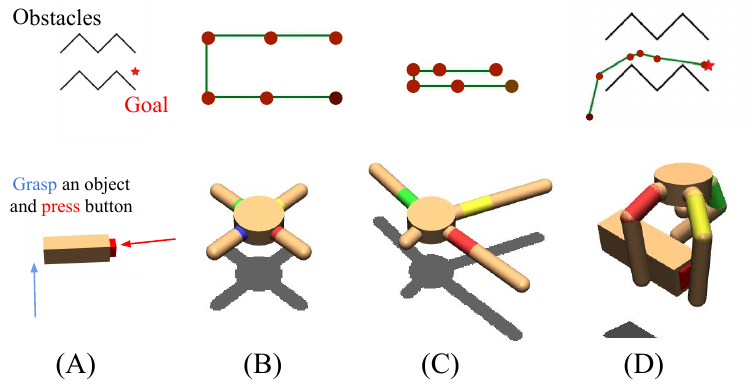}
    \caption{\OURS co-learns hardware design parameters and control policies, exemplified here on a 2D reaching task (top row) and a 3D manipulation task (bottom row). (A) shows each task. (B) shows the initial design of the robots. (C) visualizes the optimized designs resulting from \OURS. (D) shows the robot executing the control policy that has been co-optimized alongside the design parameters.}
    \label{fig:eye-candy}
    \vspace{-2em}
\end{figure}
\begin{figure*}[t]
    \centering
    \includegraphics[width=0.9\linewidth]{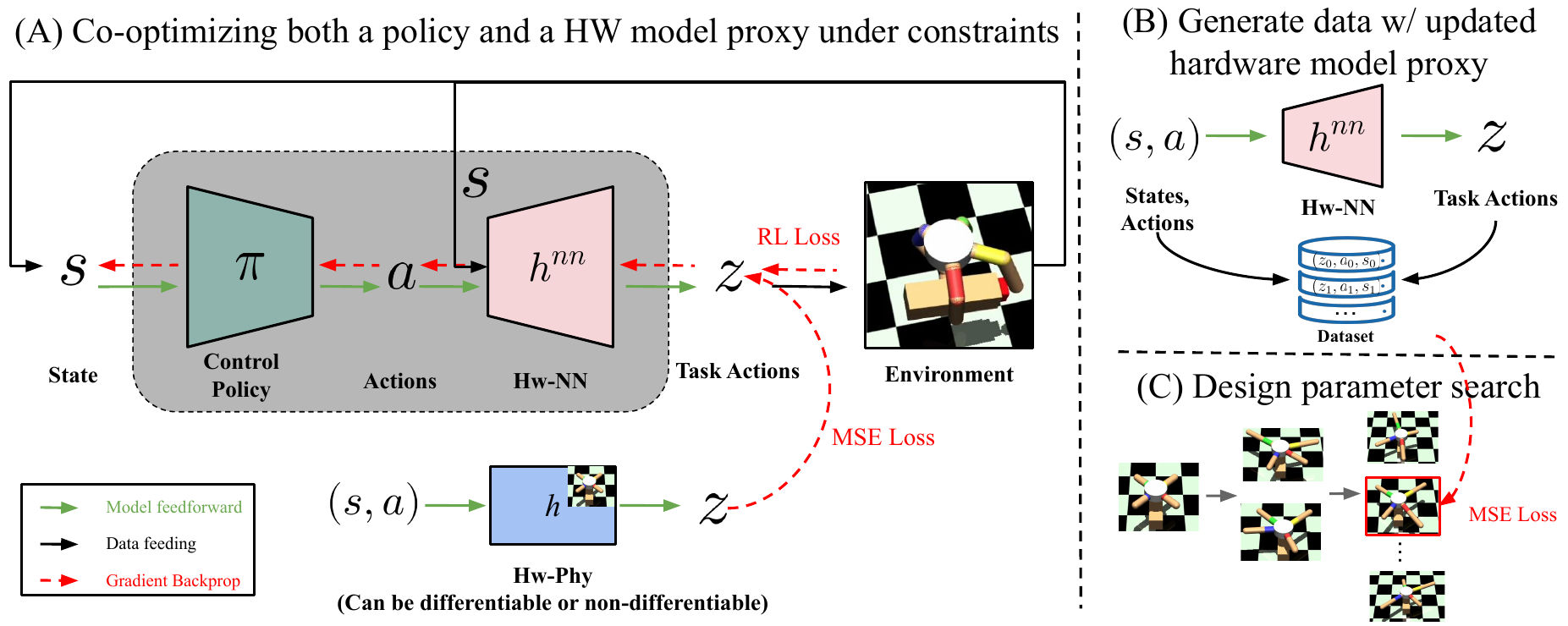}
    \caption{\textbf{Approach overview}. \OURS is an iterative training framework for design-control co-optimization: (A) We first co-optimize both a control policy and a neural-network proxy of the hardware model (Hw-NN) with an RL loss and a constraint loss computed using the more realistic, physics-based hardware model (Hw-Phy); (B) Using the updated Hw-NN, we construct a dataset $\mathcal{D}$ of tuples of states, actions, and task actions; (C) Using $\mathcal{D}$, we search for the design parameters that match Hw-Phy to the Hw-NN proxy. }
    \label{fig:method}
    \vspace{-1.5em}
\end{figure*}
In this paper, we propose to address these challenges by considering the cumulative effect of hardware design parameters on the behavior of the robot itself, rather than approaching them just as values to be optimized. With this in mind, we differentiate between two components: 
\begin{itemize}
    \item A physics-based hardware model, dubbed Hw-Phy. This is a traditional model, designed to mimic the behavior of real hardware as accurately as possible, and typically implemented by simulating some aspects of the laws of physics. Its parameters include the design parameters that are the goal of the optimization. Depending on the underlying method used, Hw-Phy may or may not be differentiable. 
    \item A neural network-based hardware model proxy, dubbed Hw-NN. The job of Hw-NN is to help with the co-optimization problem. Specifically, our method ensures that, during the optimization, Hw-NN remains as close as possible to Hw-Phy. However, owing to its implementation as a neural network, Hw-NN is always differentiable, which allows its integration into an efficient co-optimization routine, one that also optimizes a software control policy.
\end{itemize}
In the proposed framework, we use RL to co-optimize a control policy alongside the Hw-NN model, under the constraints that Hw-NN needs to mimic a real robot's behaviors as encapsulated by Hw-Phy. The advantage is that we no longer require a differentiable physics simulation in our co-optimization pipeline, but our policy still receives information about how the design parameters affect robot behavior during training. The result is an iterative training procedure: 1. Using Hw-Phy as a constraint, we optimize the control policy along with Hw-NN to maximize the task performance; 2. With the improved policy, we search for design parameters that match Hw-Phy with the current version of Hw-NN. Conceptually, the first training phase aims to find a combination of hardware and control policy that can complete the tasks. The second training phase aims to ensure that the optimized hardware is still realistic, given real-world physical constraints.

We dub our method \textbf{M}odel \textbf{O}ptimization via \textbf{R}einforcement and a \textbf{P}roxy for \textbf{H}ardware, or \textbf{\OURS}. By separating task learning and design derivation into two phases, \OURS enables the use of a non-differentiable Hw-Phy model, and can combine the ability of RL to reason about long-horizon behaviors with the use of gradient-free algorithms for parameter search. We summarize our overall contributions as follows:
\begin{compactitem}
\item We propose a novel method that co-optimizes both the design and policy of a robot directly in parameter space with RL without the assumption of the differentiability of the hardware model.
\item We propose a technique that mitigates the optimization difficulty of improving the robot's task performance while imposing realistic constraints on the hardware model.

\end{compactitem}

\section{Related Work}

Considerable research has investigated co-optimizing the design and control of a robot~\cite{LAN2021107688, Ha2017JointOO, Cheney2017ScalableCO, xu2021moghs, Xu2021AnED, Spielberg2019cosoft, ha2018coopt}. 
One approach to this is using gradient-free optimization methods and treating the evaluation of a design as a black box.
For instance, Nygaard \etal \cite{Nygaard2018realworldcoop} apply evolutionary algorithms on a quadrupedal robot to optimize its leg lengths and control parameters in the real world. Deimal \etal \cite{Deimel2017AutomatedCO} apply particle filter optimization method to optimize the shape of a soft robotic hand and grasping poses in simulation. 
Liao \etal \cite{Liao2019DataefficientLO} propose a Bayesian optimization method to tune the design of a microrobot efficiently to a walking task.
Xu \etal apply graph heuristic search \cite{xu2021moghs} with RoboGrammar \cite{zhao2020robogrammar}, which is a set of graph grammar for robot design, for terrestrial robot design.
However, gradient-free optimization methods suffer from long-horizon reasoning and can not be used for evaluating a robot with complex tasks. Our work aims to optimize the control and hardware of robots for long-horizon tasks. Therefore, we use RL for joint optimization.

Recently, RL has been considered for design and control co-optimization~\cite{jackson2021orchid, Ha2018ReinforcementLF, schaff2019jointly}.
Chen \etal \cite{chen2020hwasp} propose to model the robot as a computational graph differentiably and co-optimize its parameters and a control policy using RL. 
Wang \etal suggest Neural Graph Evolution (NGE) \cite{wang2018neural}, which models the structure of an agent as a graph neural network (GNN), optimizes the morphology of an agent by changing its graph structure, and learns to control by adapting the parameters of the GNN.
Luck \etal \cite{Luck2019DataefficientCO} propose to learn a latent space that represents the design space and train a design-conditioned policy.
Transform2Act \cite{Yuan2021Transform2ActLA} considers a transform stage when actions can modify a robot's kinematic structure and morphology, then a control stage when the design is frozen and the policy only computes control actions. 
\OURS directly optimizes design in the parameter space of a policy without assuming the differentiability of the robot by separating task optimization and design parameter search into two steps.

In this work, we observe the existence of optimization difficulty caused by gradient interference introduced from the mismatch between RL improvement and learning realistic hardware.
This is similar to the observed optimization difficulty in multi-task learning literature \cite{Hessel2018MultitaskDR, Kendall2017MultitaskLU} if we treat task improvement and being realistic as two tasks.
Previous works have explored mitigating the conflicts for multi-task learning. 
For example, Senor and Koltun \cite{Sener2018MultiTaskLA} scale the gradients introduced by different tasks to reduce the scale difference among gradients. 
GradNorm \cite{Chen2017GradNormGN} uses gradient normalization to facilitate multi-task learning.
Our work is inspired by PCGrad \cite{Yu2020GradientSF}, which uses cosine similarity between gradients to measure the conflicts and project a gradient to the normal plane of another when conflicting. In this work, we treat the task learning and hardware constraints as a dual-task learning problem and project the task learning gradients to the normal of hardware constraint gradients.

\section{Method}
\label{sec:method}

\subsection{Preliminaries}
We formulate our co-optimization problem as a Markov Decision Process (MDP). An MDP can be represented by a tuple $(\mathcal{S}, \mathcal{A}, \mathcal{F}, \mathcal{R})$, where $\mathcal{S}$ is state space, $\mathcal{A}$ is the action space, $\mathcal{R}(\bm{s}, \bm{a})$ is the reward function, and $\mathcal{F}_{\phi}(\bm{s}' | \bm{s}, \bm{a})$ is the state transition model, where $\bm{s}, \bm{s}' \in \mathcal{S}$, and $\bm{a} \in \mathcal{A}$. The transition function is parameterized by some design parameters $\phi$, which determine the behaviors of the robot. The goal of solving this MDP is finding both a control policy $\pi_{\bm{\theta}}(\bm{a}|\bm{s})$ and design parameters $\phi$ that optimize the expected returns: $\mathbb{E}[\sum_{t=0}^T\mathcal{R}(\bm{s}_t,\bm{a}_t)]$ where $T$ is the length of an episode.

The key idea of \OURS is modeling the cumulative effect of the hardware design parameters $\phi$ on the real robot and task performance. Consider the case where $\phi$ comprises the kinematic parameters of the robot (e.g. link lengths, mounting locations, geometry, etc.). Hardware essentially converts policy actions (joint angles) into task-related actions (end-effector movements). In this example, a physics-based hardware model simply consists of the forward kinematics function. In general, we refer to such a physics-based hardware model as Hw-Phy. We model its effects using a function $h()$ that, given a state, converts from policy actions $a$ to task-related actions $z$: $h=h_\phi(\bm{z} |\bm{s}, \bm{a})$.

However, instead of directly integrating Hw-Phy and its parameters into our optimization routine, we use a neural network-based proxy that we refer to as Hw-NN. As a proxy for Hw-Phy, HW-NN takes similar inputs and produces similar output, thus we model its behavior as the function $h^{nn}=h^{nn}_\psi(\bm{z} |\bm{s}, \bm{a})$. The key difference is that Hw-NN is always differentiable (as it is modeled as a neural network) and provides additional flexibility compared to Hw-Phy. 

Both $h_\phi()$ and $h^{nn}_\psi()$ are parameterized, but the nature of these parameters is vastly different. The parameters $\phi$ of $h()$ have physical meaning and correspond directly to the design parameters we wish to determine. In contrast, the parameters $\psi$ of $h^{nn}()$ are just the weights and biases of a neural network, with no physical correspondent.

By using a Hw-NN, we now can co-optimize the parameters of $h^{nn}$ with the control policy parameters. The goal of the optimization is to improve task performance, which is evaluated by expected returns, and approximate $h$ using $h^{nn}$. However, only optimizing $h^{nn}$ and a policy $\pi$ does not satisfy our goal of extracting a design that we can build in the real world since the parameters of $h^{nn}$ are not interpretable by humans. Hence, we propose to search explicit parameters $\phi$ that mimic the Hw-NN with good task performance.

Therefore, the resulting training pipeline is an iterative process (see Fig. \ref{fig:method}): We first co-optimize both the control policy and the Hw-NN to task performance, under the constraints that the Hw-NN remains close to the current version of Hw-Phy; Then, we search for the hardware design parameters that allow Hw-Phy to match the current version of the Hw-NN.

\subsection{Hardware as constraints}

The first step of our framework is to co-optimize both the control policy and the Hw-NN to improve task performance under hardware constraints.
By using a Hw-NN $h^{nn}$, we now can extend our policy to consider the effect of the design parameters: $\pi^{comb} = \pi_{\theta}(\bm{a}|\bm{s})h^{nn}_{\psi}(\bm{z}| \bm{a}, \bm{s})$.
The optimization goal of $\pi^{comb}$ is a constrained optimization problem:
\begin{align}
    &\max_{\theta, \psi}\mathbb{E}_{\pi, h^{nn}}[\sum_{t=0}^T\mathcal{R}(\bm{s}_t,\bm{z}_t)] \nonumber\\
    &\text{subject to } D[h(\bm{z}|\bm{s}, \bm{a}),  h^{nn}(\bm{z} | \bm{s}, \bm{a})] \le \epsilon \nonumber
\end{align}
Here, $D$ is a divergence function, which measures the divergence between the Hw-NN $h^{nn}$ and the Hw-Phy $h$.  $\epsilon$ is some chosen small constants. 

In practice, instead of directly performing constrained policy optimization, our work optimizes both $\pi^{comb}$ via an unconstrained objective that uses the divergence between Hw-NN and Hw-Phy as a regularization term:  
\begin{align}
    \label{eq:objective}
    &\max_{\theta, \psi}\mathbb{E}_{\pi, h^{nn}}[\sum_{t=0}^T\mathcal{R}(\bm{s}_t,\bm{z}_t) - \alpha D[h(\bm{z}|\bm{s}, \bm{a}),  h^{nn}(\bm{z} | \bm{s}, \bm{a})]]
\end{align}
where $\alpha$ is a constant.

Since the divergence between $h$ and $h^{nn}$ is intractable, we measure it via a sampling-based method. Essentially, we collect state-action pairs $(s, a)$ and feed them to both models. Then, we use the distances between outputs from both models as an estimate of the divergence between the two models. Note that in this process, only the parameters of $\pi$ and $h^{nn}$ are optimized. We use actor-critic RL in this step.

\subsection{Deriving design parameters}
Our ultimate goal is to derive design parameters from our optimization process to build a robot. However, in the policy learning step, $h^{nn}$ is a neural network whose parameters are uninterpretable by humans. Although this learning step indeed produces some policies that can achieve high task performance, its product cannot be used to build a real robot.

Hence, the second step of our co-optimization framework is to search for design parameters that match the performance of the updated Hw-NN:
\begin{align}
    \label{eq:obj2}
    &\min_{\phi}D[h_{\phi}(\bm{z}|\bm{s}, \bm{a}),  h^{nn}_{\psi}(\bm{z} | \bm{s}, \bm{a})]
\end{align}
Similar to the policy optimization step, we use a sampling-based method to estimate the divergence between $h$ and $h^{nn}$.

Here, we do not have any restriction on the differentiability of the Hw-Phy $h$. If $h$ is differentiable, we can apply gradient-based optimization methods, e.g. stochastic gradient descent, for design parameter searching. If $h$ is non-differentiable, we can use a non-differentiable optimization method, e.g. evolutionary algorithms. 

Without the need to reason long-horizon behaviors, evolutionary algorithms can search in the design space to find parameters that match the optimized Hw-NN well. In this work, we use Covariance Matrix Adaptation - Evolution Strategy (CMA-ES) \cite{hansen2001completely} for design parameter derivations.

Overall, \OURS is an iterative process that first improves both control and Hw-NN to task performance. Then, based on the adapted version of the Hw-NN, \OURS extracts design parameters that mimic the behaviors of Hw-Phy.
\vspace{-0.5em}
\begin{algorithm}
\caption{\OURS}
\label{alg:transform}
\begin{algorithmic}[1]
    \WHILE {returns have not converged}
        \STATE Sample $H$ trajectories using the current control policy $\pi_{\theta}$ and Hw-NN $h^{nn}_{\psi}$ with an MDP \\
        \STATE Optimize parameters of the control policy $\pi_{\theta}$ and Hw-NN $h^{nn}_{\psi}$ with objective E.q. \ref{eq:objective}\\
        \FOR{every K steps}
            \STATE With updated Hw-NN $g$, compute updated outputs $z$ using the current policy and construct a dataset $\mathcal{D} = [(s_{0}, a_0, z_0), (s_{1}, a_1, z_1),... ,(s_{m}, a_m, z_m)]$ \\
            \STATE Optimize Hw-Phy $h$ using data $\mathcal{D}$ with E.q. \ref{eq:obj2} \\
        \ENDFOR \\        
    \ENDWHILE
\end{algorithmic}
\end{algorithm}

\subsection{Objective mismatch between task learning and hardware constraints}
While \OURS by itself provides a method to co-optimize the design and control, its final objectives can be seen as a multi-task learning objective: the first part improves its task performance and the second part constrains the Hw-NN to be close to Hw-Phy. 
These two objectives do not necessarily agree on how to adapt the parameters.
In practice, the mismatch between these two objectives can result in detrimental gradient interference that makes optimization challenging (detailed discussion in Section \ref{sec:gradients}).

Hence, in this work, we adopt PCGrad \cite{Yu2020GradientSF} and project the RL gradients to the normal plane of the design constraint gradient direction if they conflict.
Similar to PCGrad, we measure the conflict between two gradient directions by their cosine similarity $S_c$ -- if two gradients conflict with each other, the cosine similarity of the two directions is negative:
\[
S_c(g_{task}, g_{hw}) = \frac{g_{task} \cdot g_{hw}}{||g_{task}|| \cdot ||g_{hw}||}
\]
 Here, $g_{task}$ and $g_{hw}$ represent gradients introduced by the RL loss and the divergence between $h$ and $h^{nn}$ accordingly.

If the cosine similarity is negative, we project task gradients $g_{task}$ to the normal plane of hardware gradients $g_{hw}$:
\[
g_{task} = g_{task} - \frac{g_{task} \cdot g_{hw}}{||g_{hw}||}\cdot g_{hw}
\]
This projection prioritizes the learning of the hardware proxy model. This is crucial for hardware-policy co-optimization since policy gradients can be misleading if $h^{nn}$ does not model the effect of actions and design parameters well.

\section{Experiments}
\label{sec:experiments}
\begin{figure}
    \centering
    \includegraphics[width=\linewidth]{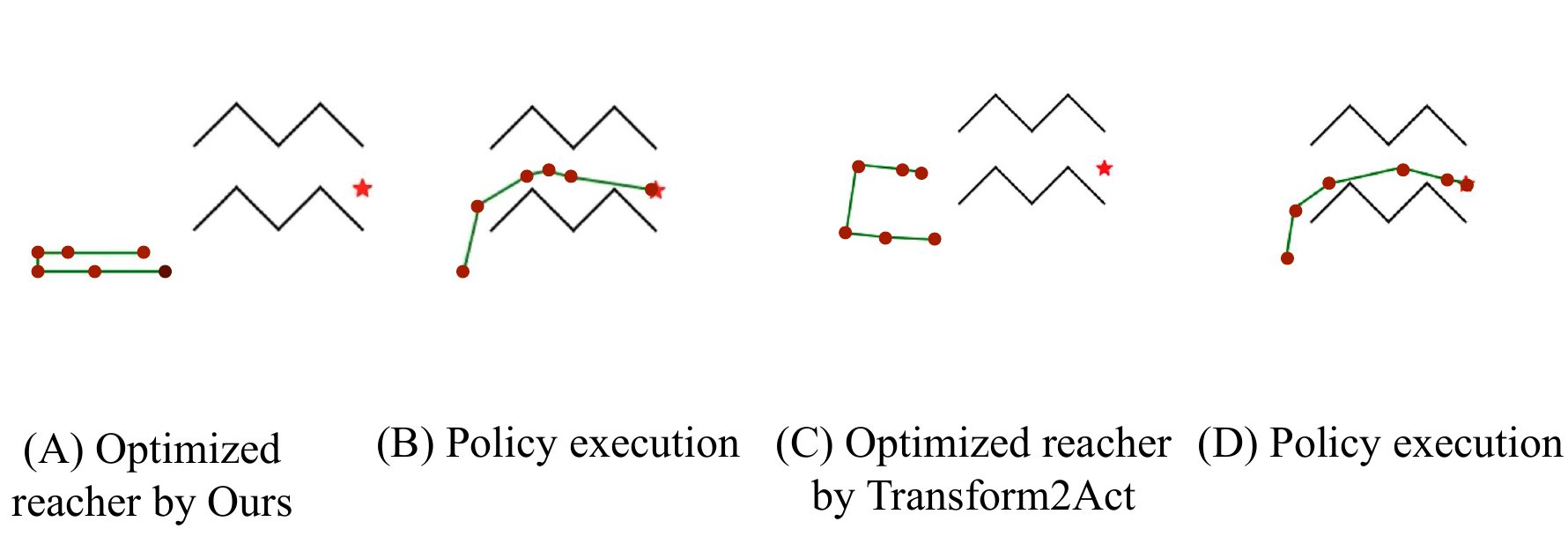}
    \caption{\textbf{Optimizing a reaching robot.} (A) and (C) show the optimized reacher by \OURS and \TRANSFORM respectively. (B) and (D) visualize the co-optimized control policies for \OURS and \TRANSFORM respectively. Both methods are able to optimize the link lengths to complete the reaching task.}
    \label{fig:reach2}
    \vspace{-0.5em}
\end{figure}
\begin{figure}
    \centering
    \includegraphics[width=\linewidth]{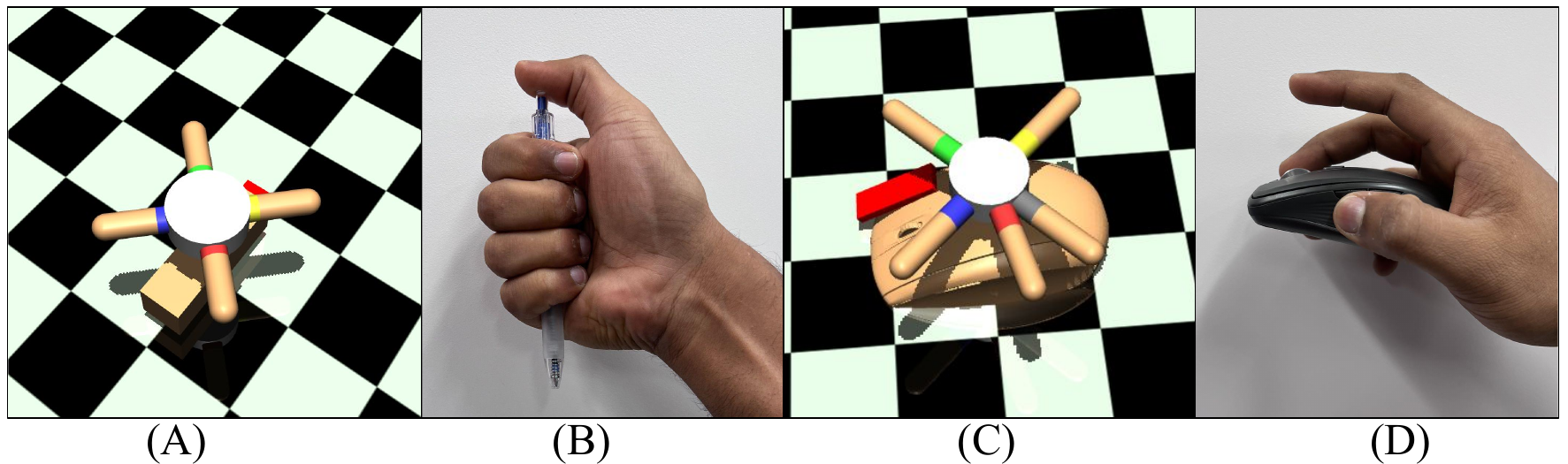}
    \caption{\textbf{Button pressing tasks.} (A) pen-pressing task where the button is located at one end of a long object. (B) pen-pressing as performed by a person. (C) mouse-clicking task where the button is located at a corner of the object. (D) mouse-clicking task as performed by a person. }
    \label{fig:hand-task}
    \vspace{-1.3em}
\end{figure}
\OURS aims to find robust hardware design parameters and control policies that can solve complex robotic problems. Therefore, we test our algorithm on problems with the following characteristics: 1. Only a small part of the design space allows the robot to complete the task; 2. The final goal of our tasks is hard to explore in the task environment. 
We evaluate \OURS with task performances using Hw-Phy and compare it against the performance of unoptimized hardware.

One of the key characteristics of our method is that \OURS optimizes both the design and policy in parameter space. To evaluate its performance, we compare our method with three baselines: 
\begin{compactitem}
    \item \textbf{Transform2Act}~\cite{Yuan2021Transform2ActLA}: This approach treats design as actions and uses Proximal Policy Optimization (PPO) to find good design and control parameters.
    \item \textbf{CMA-ES with RL inner-loop}: This approach uses CMA-ES for searching hardware parameters and trains a policy for control. The negative best return achieved by the RL policy is used as a cost for CMA-ES. 
    \item \textbf{RL-NoHWOpt}: This approach trains an RL agent using the initial design parameters and do not optimize the hardware parameters of the robot.
\end{compactitem}

\subsection{Optimizing a reaching Robot}

We first test our method on a 2D reaching task. 
Here, we require a 5-link reaching robot to navigate a zig-zag-shaped tunnel and touch a goal location with the end-effector.
The design optimization goal is finding the appropriate link length for each length so the robot can reach the goal with minimal collisions. 
Overly long links can make moving in the constrained space without collision difficult, and overly short links may decrease the possibility of exploring the goal. 
The optimization range of the link lengths is $[0.05, 5]$ and the initial design of the robot has a link length of $3$.

To optimize link lengths, it is sufficient to choose forward kinematics as the Hw-Phy, which converts joint space actions and current joint states to the end-effector space: $h = h(a_{e.e.}|a_{joint}, s)$, where $a_{joint}$ is the output from $\pi$ and represents the change of joint positions. 
The state space contains joint positions and end-effector positions. 
This problem space is difficult to explore since the desired behavior requires navigating tunnel. Hence, we use a reward function that encourages exploring inside the tunnel:
\begin{align}
    R &= ||p_{e.e.} - p_{goal}|| + \beta_{0}||y_{e.e.} - y_{tunnel}|| \nonumber \\
    & + \beta_{1} r_{collide} + \beta_{2}r_{goal}\nonumber
\end{align}
Here, $p_{e.e.}$ and $p_{goal}$ and the Cartesian coordinate of the end-effector and goal accordingly. $y_{e.e.}$ and $y_{tunnel}$ represent the $y$ coordinate of the e.e. and the center of the tunnel. 
$r_{collide}$ is a collision penalty. 
$r_{goal}$ rewards the agent touch the goal location with its e.e.. $\beta_{0}, \beta_{1}, \beta_{2}$ are constants.

\subsection{Optimizing robotic hands with manipulation tasks}
\begin{figure}
    \centering
    \includegraphics[width=\linewidth]{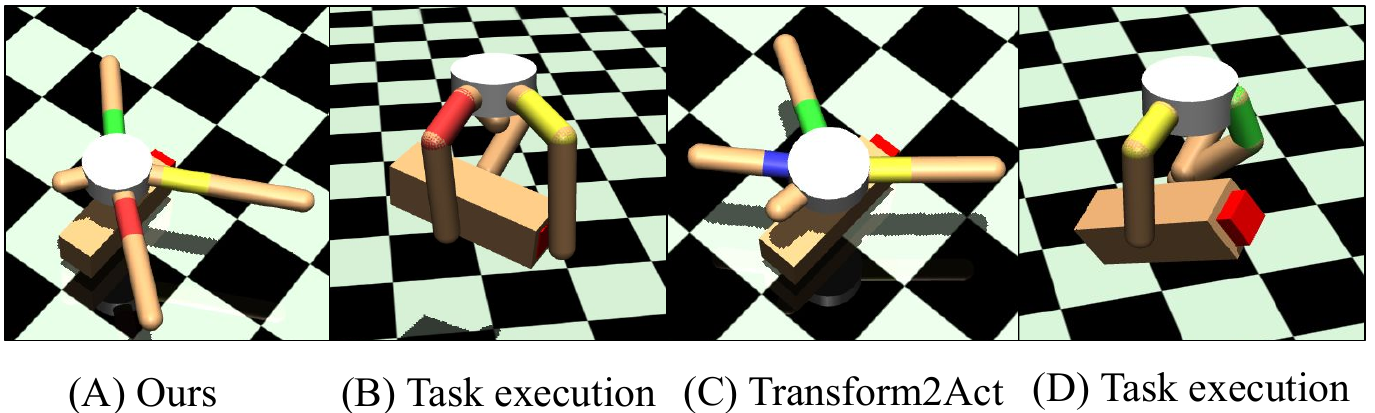}
    \caption{\textbf{Optimized hand with pen-pressing}. (A) shows the optimized robotic hand using \OURS. (B) shows the optimized hand completing the pen-pressing task. (C) shows an optimized hand using \TRANSFORM. (D) shows the \TRANSFORM agent executing its control policy and failing to press the button.}
    \label{fig:handopt1}
    \vspace{-1.5em}
\end{figure}

In this experiment, we use \OURS to optimize the kinematics of robotic hands for complete object manipulation tasks. 
As shown in Fig. \ref{fig:hand-task}, the robot needs to grasp the object and press a button on the object in hand.
This task is challenging for several reasons: 1. The agent needs to find a design that produces a stable grasp so it can press the button without dropping the object; 2. The goal of pressing a button in hand can only be discovered after a stable grasp of the object.

We optimize the angle of the finger placement and the link length of the finger links.
Each finger has two movable links and palm movement is constrained to the z-axis. In this case, the robot can not complete the button-pressing task by adjusting its grasp pose.
Hence, the optimization algorithm needs to find a suitable finger placement to complete the task. In this task, the optimization range of finger placement angles is $[-\pi, \pi]$ and the range of link lengths is $[0.02, 0.4]$. In this experiment, we assume that two links of a finger share the same link length.

Unlike the reaching task, it is unclear how to define a task-related action space for a contact-rich task. Hence, for the button-pressing tasks, we use the full transition function $h = \mathcal{F}(s'| s, a)$ as the Hw-Phy. In practice, we use the physics simulator Mujoco \cite{mujoco} as the transition function. The state space of this task contains joint positions of the hand joints, the location of the robotic hand, and the pose of the object. The reward function for this task is:
\[
R = ||p_{obj} - p_{goal}|| + \beta_{0}r_{contact} + \beta_{1}z_{obj} + \beta_{2}r_{goal}
\]
Here, $p_{obj}$ and $p_{goal}$ are the location of the object and the goal accordingly. $r_{contact}$ is $1$ if the distal link of any finger makes contact with the button. $z_{obj}$ represents the height of the object. $r_{goal}$ is a reward bonus only supplied when the robot presses a button with one of its distal links and the object is in-hand. $\beta_{0}, \beta_{1}, \beta_{2}$ are constants.

We test our algorithm to learn hardware parameters to manipulate two buttoned objects: 
1. Pen-pressing: In this task, the button is located on one end of a long box object. This is similar to pressing a push button on a pen. 2. Mouse-clicking: In this task, the button is located on the corner of a computer mouse. The mouse geometry is more complex and requires more fingers in contact to form a stable grasp. As shown in Fig. \ref{fig:hand-task}, both objects are designed to be used by human hands with different grasp poses.

\section{Results}
\label{sec:results}
\begin{figure}
    \centering
    \includegraphics[width=\linewidth]{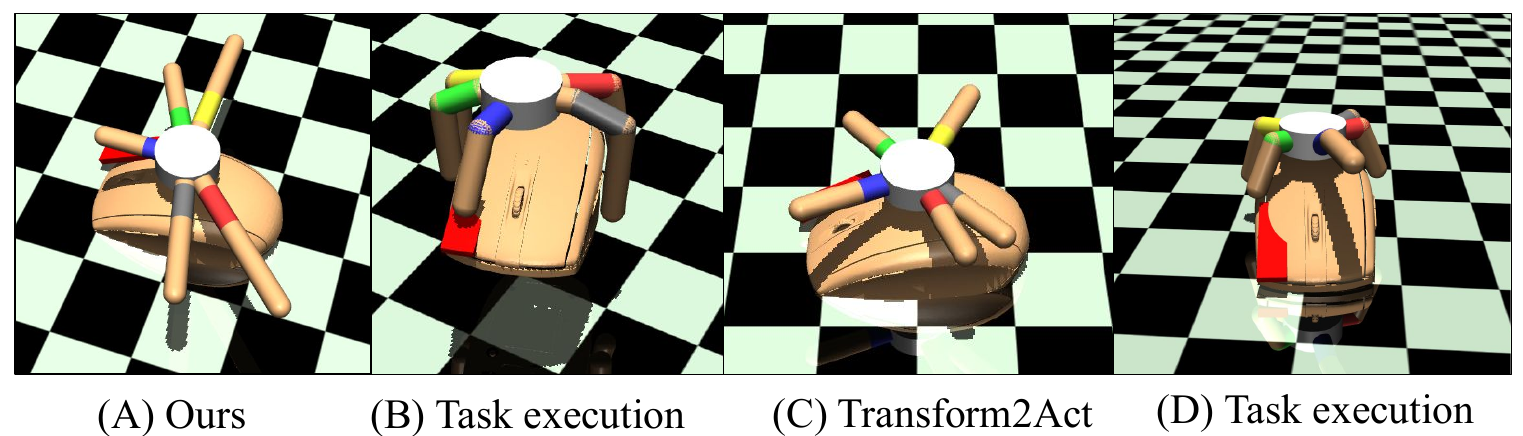}
    \caption{\textbf{Optimized hand with mouse-clicking}. (A) and (C) show the optimized robotic hand using \OURS and \TRANSFORM respectively. (B) shows the \OURS hand completing the mouse-clicking task. (D) shows the \TRANSFORM agent failing to lift the computer mouse with short fingers.}
    \label{fig:handopt2}
    \vspace{-1.2em}
\end{figure}

\begin{figure}
    \centering
    \includegraphics[width=0.9\linewidth]{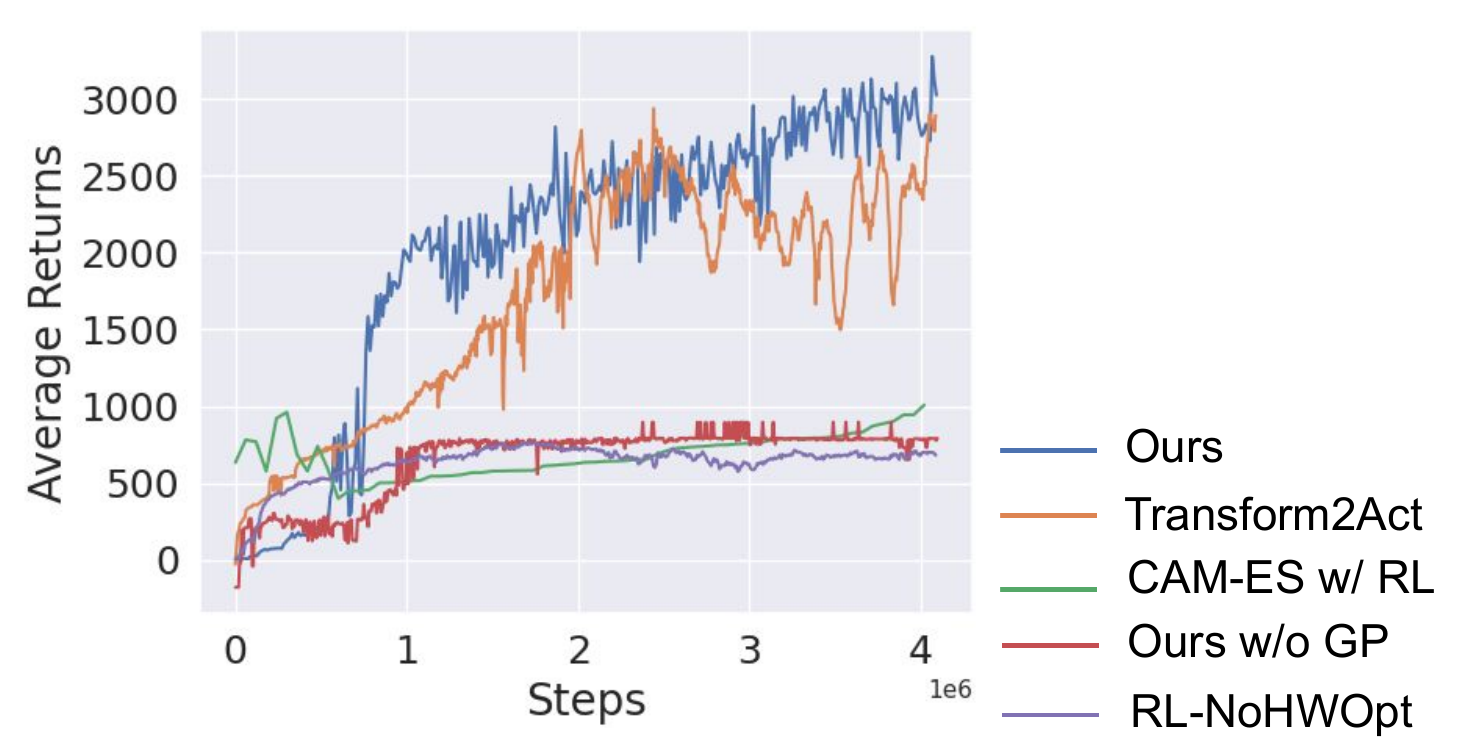}
    \caption{\textbf{Average returns vs. environment steps for the reaching task}. Here, GP represents gradient projection.}
    \label{fig:reach_returns}
    \vspace{-1.5em}
\end{figure}

\begin{figure}[t]
    \centering
    \includegraphics[width=\linewidth]{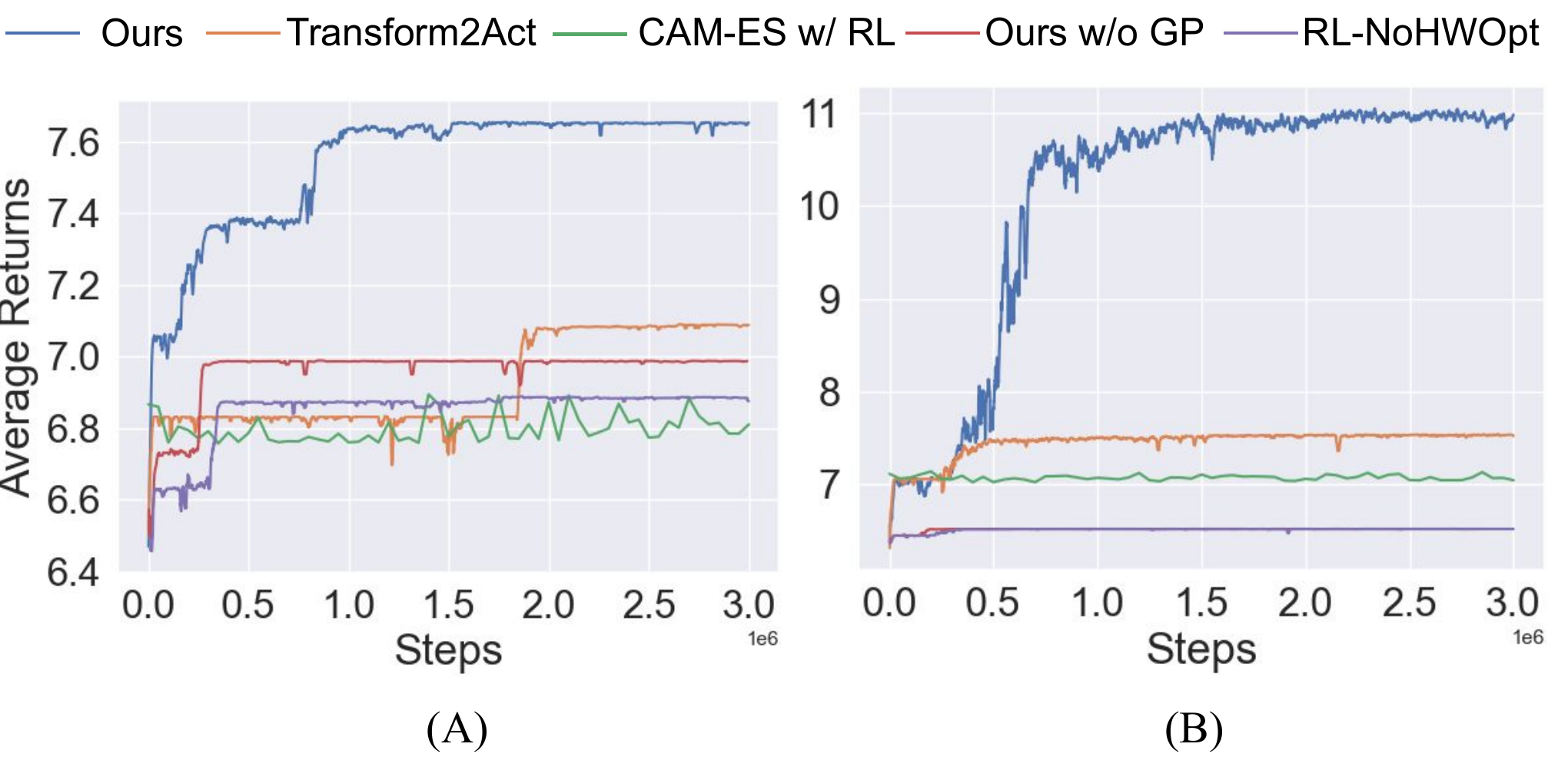}
    \caption{\textbf{Average returns (in log scale) vs environment steps for the button pressing tasks}. (A) pen-pressing; (B) mouse-clicking.}
    \label{fig:comb_returns}
    \vspace{-1em}
\end{figure}

\subsection{Task performance}
Our results show that \OURS can learn control policies and designs that achieve good task performance for the reaching task. 
As shown in Fig. \ref{fig:reach2}, the robot shortens the lengths of the third and fourth links to decrease collisions when navigating in the tunnel.  The final optimized link lengths are $\{2.51, 2.10,  0.66,  1.01, 2.3\}$.
On the other hand, \NOHW cannot discover the goal since it collides with the tunnel walls. This indicates that the original design is not suitable for learning this task.
As shown in Fig. \ref{fig:reach_returns}, Transform2Act also achieves similar task performance as \OURS. It successfully finds hardware parameters that allow it to discover the goal and finally converge to a hardware design that can complete the task. 
Finally, CMA-ES with RL inner-loop fails to find good design parameters to achieve the goal in a similar timescale.

For the pen-pressing task, \OURS discovers behaviors that use only two fingers to grasp the object and elongate one finger to press the button (Fig. \ref{fig:handopt1}). 
It also shortens the unused finger to avoid any contact that results in unstable grasps.
The final optimized link lengths are $\{0.11, 0.24, 0.33, 0.30\}$ and the final finger placement differences from the initial design are $\{0.16, -0.06, 0.32, 0.07\}$ radians.
As shown in Fig. \ref{fig:comb_returns}, \OURS outperforms all the baselines in the pen-pressing task. Both \NOHW and \CMAES fail to learn to grasp the object stably. \TRANSFORM learns a design that can grasp the object. However, it fails to discover the final goal of pressing the button. 

Finally, for the mouse-clicking task, \OURS is able to elongate a robot's links to establish a stable grasp of the computer mouse and click the button.
The optimized link lengths are $\{0.17, 0.24, 0.27, 0.3, 0.28\}$ and the optimized finger placement angle differences are $\{0.78, 0.32, -0.7, 0.54, -0.01\}$ radians.
For the baselines, all of them fail to achieve a stable grasp. \NOHW has difficulty grasping the object using short fingers. Both \CMAES and \TRANSFORM fail to find appropriate link lengths and finger positions that lift the object stably.

Our results for co-optimization demonstrate that \OURS is able to optimize robots that learn different tasks. 
Crucially, \OURS efficiently explores the design space and is able to discover hardware design parameters that allow for rich explorations in the task environment. Compared to the baselines, which can only learn part of the manipulation task, \OURS can discover the final goal from a delayed task-completion reward signal.

\subsection{Discussion}
\label{sec:gradients}
\begin{figure}
    \centering
    \includegraphics[width=\linewidth]{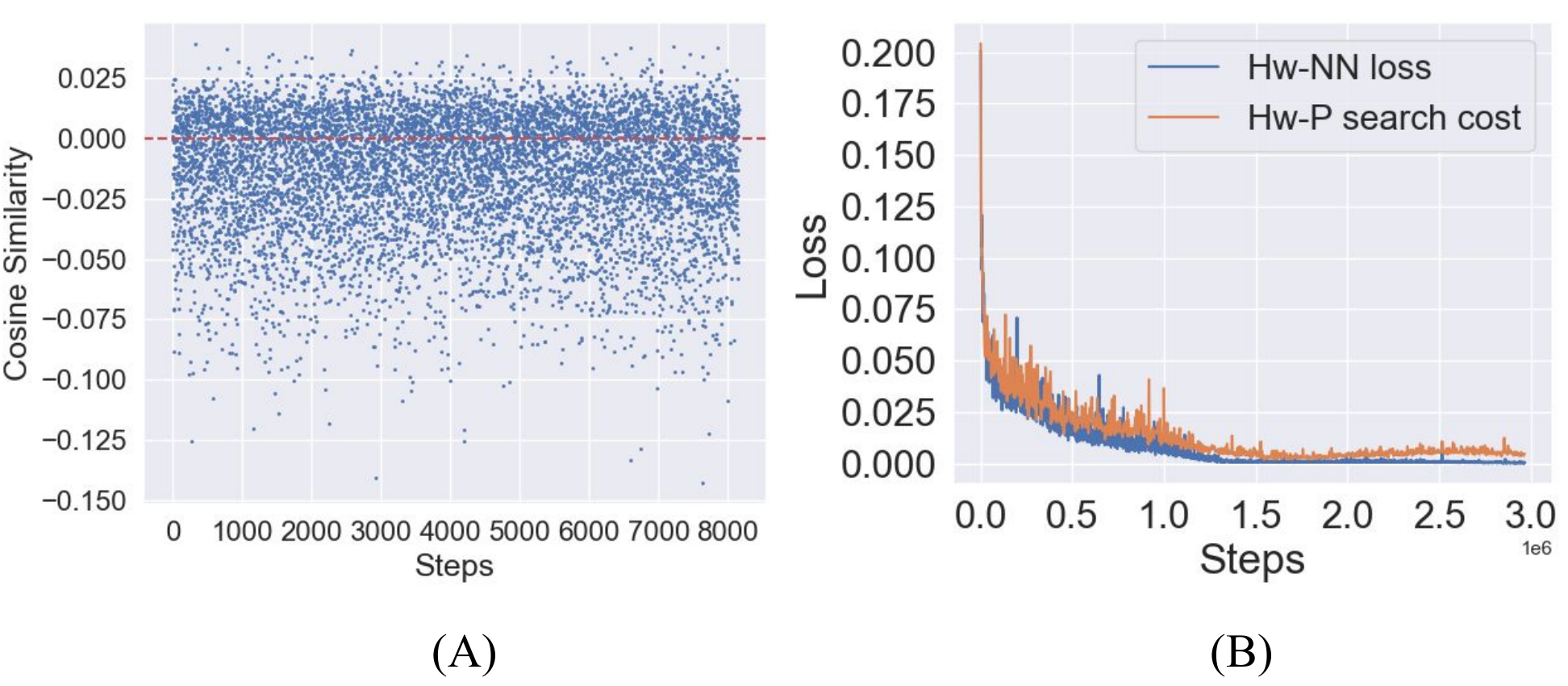}
    \caption{\textbf{Training details for mouse-clicking}. (A) shows cosine similarity between hardware constraint gradients and RL gradients. (B) plots constraint loss of Hw-NN and search costs for Hw-Phy.}
    \vspace{-1em}
    \label{fig:training_details}
\end{figure}

As shown in Figs. \ref{fig:reach_returns} and \ref{fig:comb_returns}, \OURS fails to learn the task without using gradient projection, which implies that gradient projection is crucial to learning hardware and a control policy with high task performance. To further investigate this, we visualize the cosine similarity between the RL gradients and hardware proxy learning gradients in Fig. \ref{fig:training_details}. The cosine similarity is negative for approx. $64\%$ of the training steps, meaning that the learning progress for both task improvement and hardware approximation can often be hindered by gradient interference. Hence, gradient projection is a critical component.

Our training framework relies on the Hw-NN to learn a policy from the cumulative effect of design parameters. 
If the Hw-NN is inaccurate, the control policy may never be learned since the policy gradients can be misleading. 
Hence, it is important that the Hw-NN is close to the Hw-Phy (i.e. minimizing $D[h_{\phi}(\bm{z}|\bm{s}, \bm{a}),  h^{nn}_{\psi}(\bm{z} | \bm{s}, \bm{a})]$). 
Since our framework is iterative, this minimization is achieved from two directions: the Hw-NN mimics the behaviors of an Hw-Phy, and vice versa.
As shown in Fig. \ref{fig:training_details}, both Hw-NN's loss and Hw-Phy's search costs are high at the beginning of the training. While a high initial loss for Hw-NN is expected, the search cost is also high because, in this stage, the hardware proxy model is close to random and it is hard to find realistic parameters that allow Hw-Phy to match it. However, both the loss and the cost decrease during training, and finally both models converge to to each other.

\section{Conclusion and Future Directions}
\label{sec:conclusion}

We introduced \OURS, a method to co-optimize hardware design parameters and control policies. 
\OURS uses a hardware proxy model, Hw-NN, that learns the cumulative effect of hardware design parameters on the robot itself. With Hw-NN, our method does not require a differentiable physics model to compute gradients of the design parameters. 
Our results show that \OURS can learn hardware design parameters and control policies that enable hard-exploration manipulation tasks. 

A key limitation of our approach is that it currently does not learn the morphology of the robot, which is assumed as given. However, we envision that \OURS can achieve this by using a graph neural network (GNN) to model the robot, and optimizing the graph structure in order to optimize the robot's morphology. Another direction will be applying \OURS to optimize a robot design for multiple tasks, such as a robotic hand capable of diverse manipulation skills.


\clearpage
\printbibliography

\addtolength{\textheight}{-12cm}   
\end{document}